\documentclass{article}

\usepackage[preprint]{neurips_2025}

\usepackage[utf8]{inputenc}
\usepackage[T1]{fontenc}

\usepackage{amsmath}
\usepackage{amssymb}
\usepackage{amsfonts}
\usepackage{booktabs}
\usepackage{makecell}
\usepackage{multirow}
\usepackage{nicefrac}
\usepackage{microtype}
\usepackage{xcolor}
\usepackage{graphicx}
\usepackage{float}
\usepackage{threeparttable}
\usepackage{tikz}
\usetikzlibrary{arrows.meta,fit,positioning}
\usepackage{url}
\usepackage{xspace}
\usepackage[
  colorlinks=true,
  allcolors=blue
]{hyperref}

\newcommand{\method}{MAViE\xspace}
\newcommand{\hmf}{\textsc{HMF}\xspace}
\newcommand{\qatr}{\textsc{QATR}\xspace}
\newcommand{\R}{\mathbb{R}}
\newcommand{\cmark}{\textcolor{blue!65!black}{\(\checkmark\)}}
\newcommand{\best}[1]{\textbf{#1}}

\title{\method: A Multi-scale Adaptive Vision Encoder for\\
Fine-grained Visual Perception and Efficient Multimodal Reasoning}

\author{Shaofei Lei}

\begin{document}

\maketitle

\begin{abstract}
Vision-language models commonly project all tokens produced by a pretrained
vision encoder into a large language model. However, final-layer features can
discard text, local attributes, and spatial relationships, while
high-resolution inputs substantially increase context length and inference
latency. We introduce \method, a Multi-scale Adaptive Vision Encoder. \method
uses position-dependent gates to fuse shallow, intermediate, and deep features
from a vision Transformer, preserving global semantics while enhancing edges,
text, and local structure. It then performs question-conditioned token routing
according to question relevance, local information content, global semantics,
and spatial coverage, with a token budget that adapts to image complexity. To
mitigate compression loss, we further introduce full-to-compressed
representation distillation and a spatial diversity regularizer. In an
illustrative simulation under a unified 7B language-model framework, \method
reduces the average number of SigLIP-SO400M visual tokens from 729 to 146
(approximately 80.0\%) and improves the mean score on VQAv2, GQA, TextVQA,
ScienceQA-IMG, and MMBench by 2.2 percentage points, while reducing
single-image time to first token from 228\,ms to 129\,ms. We provide the full
model design and evaluation protocol. All reported numbers currently serve
only as placeholders for paper organization and experimental design; formal
claims require real training runs, independent replications, and official
benchmark evaluation.
\end{abstract}

\section{Introduction}
\label{sec:introduction}

Vision-language models (VLMs) typically consist of a vision encoder, a
vision-language connector, and a large language model. The vision encoder is
the model's gateway to the external world: once object attributes, text,
spatial locations, or local relationships are lost during encoding, even a
powerful downstream language model cannot reliably recover the missing
evidence. Work such as LLaVA~\citep{liu2023llava} has shown that connecting a
pretrained vision encoder to a language model and applying visual instruction
tuning provides a simple and scalable multimodal modeling paradigm.

Modern VLMs widely use Vision Transformer (ViT)~\citep{dosovitskiy2021vit}
encoders pretrained as CLIP~\citep{radford2021clip},
SigLIP~\citep{zhai2023siglip}, or DINOv2~\citep{oquab2024dinov2}. A common
design takes only the final encoder layer and projects every visual token
through a linear layer or a two-layer MLP. This design presents two related
challenges. First, shallow ViT layers emphasize edges, textures, and character
strokes; intermediate layers capture parts and regional structure; and deep
layers favor global semantics. Using only final-layer features can therefore
weaken OCR, counting, and local attribute recognition. Second, increasing
input resolution improves fine-grained perception but rapidly expands the
token sequence. With a fixed patch size, increasing the resolution from
\(224\times224\) to \(448\times448\) produces approximately four times as many
tokens and increases both visual self-attention and language-model prefill
cost.

Existing compression methods use spatial pooling, fixed-query resampling,
similarity-based merging, or dynamic pruning. Average pooling mixes adjacent
objects, fixed queries are independent of the current question, and
independent Top-\(K\) pruning can concentrate on only a few salient regions.
TokenPacker~\citep{li2024tokenpacker} demonstrates that visual tokens contain
substantial redundancy, but also suggests that compression must preserve both
global semantics and local evidence.

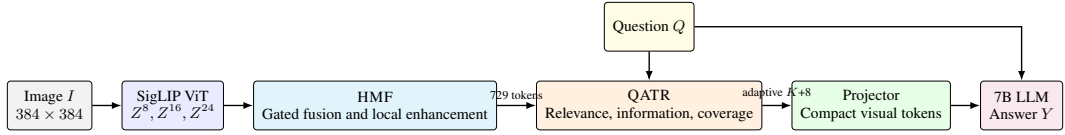
\begin{figure}[t]
  \centering
  \resizebox{\linewidth}{!}{%
  \begin{tikzpicture}[
    font=\small,
    box/.style={draw,rounded corners=2pt,minimum height=10mm,align=center,inner xsep=5pt},
    arr/.style={-{Latex[length=2mm]},thick},
    note/.style={font=\scriptsize,align=center}
  ]
    \node[box,fill=gray!10] (img) {Image \(I\)\\\(384\times384\)};
    \node[box,fill=blue!8,right=6mm of img] (vit) {SigLIP ViT\\\(Z^8,Z^{16},Z^{24}\)};
    \node[box,fill=cyan!10,right=6mm of vit] (hmf) {\hmf\\Gated fusion and local enhancement};
    \node[box,fill=orange!12,right=8mm of hmf] (router) {\qatr\\Relevance, information, coverage};
    \node[box,fill=green!10,right=6mm of router] (proj) {Projector\\Compact visual tokens};
    \node[box,fill=purple!9,right=6mm of proj] (llm) {7B LLM\\Answer \(Y\)};
    \node[box,fill=yellow!12,above=6mm of router] (question) {Question \(Q\)};
    \draw[arr] (img) -- (vit);
    \draw[arr] (vit) -- (hmf);
    \draw[arr] (hmf) -- node[above,note]{729 tokens} (router);
    \draw[arr] (question) -- (router);
    \draw[arr] (router) -- node[above,note]{adaptive \(K\)+8} (proj);
    \draw[arr] (proj) -- (llm);
    \draw[arr] (question.east) -| (llm.north);
  \end{tikzpicture}}
  \caption{Overview of \method. Multi-level features are fused by \hmf, after
  which \qatr selects an adaptive number of visual tokens conditioned on the
  image and question. Pruned tokens are compressed into eight context tokens.}
  \label{fig:overview}
\end{figure}

We propose \method to jointly address insufficient fine-grained
representations and visual-token redundancy. Figure~\ref{fig:overview}
summarizes the pipeline. First, a Hierarchical Multi-level Fusion module
(\hmf) combines shallow, intermediate, and deep representations using
position-dependent gates and preserves high-frequency details through local
enhancement. Second, a Question-conditioned Adaptive Token Routing module
(\qatr) selects tokens using question relevance, local information content,
global semantics, and spatial coverage. Finally, a complexity estimator
dynamically sets the token budget between \(72\) and \(288\), while unselected
tokens are aggregated into a small set of context tokens. Our contributions
are:
\begin{itemize}
  \item We introduce \method, a multi-scale adaptive vision encoder that
  jointly supports fine-grained perception and compact visual outputs.
  \item We propose position-dependent hierarchical gated fusion and dynamic
  token routing jointly driven by question semantics, visual complexity, and
  spatial coverage.
  \item We develop full-to-compressed representation distillation and a
  spatial diversity constraint to reduce information loss and excessive
  regional concentration.
  \item Under a unified evaluation protocol, we analyze the trade-off among
  task performance, token count, latency, and memory, while explicitly
  separating illustrative results from experiments that remain to be run.
\end{itemize}

\section{Related Work}
\label{sec:related}

\paragraph{Vision encoders.}
ViT divides an image into fixed-size patches and models global relationships
as a sequence~\citep{dosovitskiy2021vit}. Feature Pyramid Networks established
the value of semantically strong multi-scale features~\citep{lin2017fpn}, and
Swin Transformer introduced hierarchical shifted-window attention with linear
complexity in image size~\citep{liu2021swin}. NaViT further supports native
aspect ratios and resolutions through sequence packing
~\citep{dehghani2023navit}. CLIP learns a shared semantic space through
large-scale image-text contrastive learning~\citep{radford2021clip}.
SigLIP replaces globally normalized softmax contrastive learning with a
pairwise sigmoid loss~\citep{zhai2023siglip}, while DINOv2 learns transferable
image-level and pixel-level representations through self-supervision
~\citep{oquab2024dinov2}. InternVL~\citep{chen2024internvl} further scales the
vision foundation model and progressively aligns it with language. Unlike the
common practice of using only final-layer outputs, \method explicitly
preserves complementary information across levels.

\paragraph{Vision-language models.}
LLaVA~\citep{liu2023llava} connects CLIP features to a language model through a
projector and performs end-to-end tuning with visual instruction data.
Flamingo uses a resampler and gated cross-attention for interleaved visual-text
inputs~\citep{alayrac2022flamingo}, whereas BLIP-2 and InstructBLIP use
querying transformers to extract a compact, optionally instruction-aware set
of visual features~\citep{li2023blip2,dai2023instructblip}. Honeybee emphasizes
locality preservation and flexible output length in the projector
~\citep{cha2024honeybee}. Qwen-VL extends general-purpose models with grounding
and text reading~\citep{bai2023qwenvl}; Qwen2-VL introduces native dynamic
resolution~\citep{wang2024qwen2vl}, and LLaVA-OneVision unifies single-image,
multi-image, and video transfer~\citep{li2024llavaonevision}. Across these
architectures, too few visual tokens create an information bottleneck, whereas
too many increase context length, memory consumption, and time to first token.

\paragraph{Visual-token compression and efficient encoding.}
Spatial pooling is inexpensive but blurs boundaries. Fixed-query resampling
provides a stable output length but usually does not change with the question.
TokenLearner adaptively extracts a small set of visual tokens
~\citep{ryoo2021tokenlearner}; DynamicViT, A-ViT, and AdaViT progressively
reduce input-dependent computation within vision transformers
~\citep{rao2021dynamicvit,yin2022avit,meng2022adavit}. Token Merging combines
similar tokens without requiring pruning~\citep{bolya2023tome}. For
vision-language inference, PruMerge clusters redundant visual tokens
~\citep{shang2024llavaprumerge}, PAR makes reduction prompt-aware
~\citep{liu2024par}, SparseVLM performs training-free text-guided
sparsification~\citep{zhang2025sparsevlm}, and CrossGET uses cross-modal token
matching~\citep{shi2024crossget}. TokenPacker~\citep{li2024tokenpacker}
compresses through coarse-to-fine information injection, while
FastVLM~\citep{vasu2025fastvlm} reduces time to first token with an efficient
high-resolution encoder. \method additionally models image complexity and
spatial diversity and recycles information from pruned tokens.

\paragraph{Fine-grained and high-resolution perception.}
Earlier remote-sensing segmentation studies strengthen dense prediction by
deep feature fusion, frequency decoupling, and architectures specialized for
ultra-high-resolution imagery~\citep{shan2021densenet,shan2021decouple,
shan2021uhrsnet}. Global--local interaction and multi-resolution branches
further preserve context without discarding fine structures
~\citep{li2021global,shan2022mbnet}, while multilevel compactness and dispersion
improve the separability of dense features~\citep{shan2023boosting}. This
literature motivates \method's explicit preservation of shallow detail,
intermediate structure, and deep semantics. In multimodal modeling, LLaVA-UHD
similarly combines native-resolution slicing, visual-token compression, and
explicit spatial organization~\citep{xu2024llavauhd}.

\paragraph{Adaptive architectures for dense vision.}
Learnable patch proposals and dynamic refinement allocate computation
according to image content~\citep{shan2023patch,ji2024ldnet}. Recent
high-resolution networks combine self- and cross-attention, dual branches, or
asymmetric state-space and convolutional pathways
~\citep{meng2025dlnet,du2025transform,zhang2025mambacnn}. Related work also
studies global--local cross-attention, dual-relation distillation, and binary
quantization~\citep{yi2025globalcross,li2025dualrelation,ding2024binary}.
Whereas these methods adapt dense-prediction backbones, \method adapts the
visual sequence delivered to a language model.

\paragraph{Continual and data-efficient segmentation.}
Class-incremental segmentation has been approached through multi-aspect
distillation, pixel-level feature generation, and dynamic queries
~\citep{shan2021classincallaspects,shan2022classincremental,
wu2023dynamicquery}. Energy regularization and recursive least-squares fusion
offer complementary mechanisms for retaining or combining knowledge
~\citep{li2021energy,li2021fusing}. Few-shot and lifelong settings further
benefit from language-driven classifiers, contrastive selective forgetting,
and explicit organization of latent background classes
~\citep{shan2023incrementalfewshot,shan2024lifelong,shan2024organizing}.
Class-balanced active learning and flexible annotated-data generation address
the acquisition side of efficient learning~\citep{shan2024active,
yige2025flexdataset}.

\paragraph{Adaptive multimodal reasoning.}
Dynamic-resolution vision-language models directly connect input allocation
to autonomous-driving perception~\citep{zhou2025dynrsl}, and multimodal
remote-sensing interpretation similarly couples dynamic resolution with
multi-scale vision-language alignment~\citep{zhang2025multimodalrs}.
Efficiency can also be improved after visual encoding through gated,
chunked key--value caching~\citep{xu2025kvefficient}. Complementary studies
investigate cognitive memory, supervised fine-tuning for visual geolocation,
and language-aligned 3D perception~\citep{shan2025cognitive,yi2025geolocsft,
liu2025neurovoxel}. Chain-of-thought training, graph recommendation, and
cross-modal retrieval provide broader evidence that structured reasoning and
alignment mechanisms can improve multimodal decision making
~\citep{xie2025recllmr1,luo2025llmcot,sun2025gmm}.

\paragraph{Broader visual applications.}
The need to preserve localized evidence also appears in unregistered
bi-temporal change detection and trimap-free image matting
~\citep{zhao2023changedetection,shan2024matting}; in medical and scene
understanding, contrastive learning has been applied to synthetic lung
radiographs, dental X-ray segmentation, and scene-text detection
~\citep{pi2025lung,du2025dental,huang2025scene}. Finally, efficient 3D Gaussian
splatting uses geometry-guided initialization, adaptive resolution, and
density control~\citep{wang2025gdgs,liu2025adaptivegaussian}. These
domain-specific results reinforce a common principle behind \method: preserve
task-relevant fine structure while controlling representational cost.

\section{Method}
\label{sec:method}

\subsection{Problem Formulation and Architecture}

Given an image \(I\) and a question \(Q\), we resize the image to
\(384\times384\) and encode it with SigLIP-SO400M/14 using
\(14\times14\) patches. Excluding the class token, the visual sequence length
is \(N=27\times27=729\):
\begin{align}
  Z^0 &= \operatorname{PatchEmbed}(I)+P,\\
  Z^l &= \operatorname{Block}_l(Z^{l-1}),\qquad l=1,\ldots,L,
\end{align}
where \(P\) denotes the two-dimensional positional encoding. We extract
representations from layers 8, 16, and 24:
\begin{equation}
  F_s=Z^8,\qquad F_m=Z^{16},\qquad F_d=Z^{24},
\end{equation}
These correspond to shallow detail, intermediate structure, and deep
semantics, respectively. \hmf produces an enhanced representation
\(F\in\R^{N\times d}\), and \qatr selects an index set \(\mathcal{S}\), where
\(|\mathcal{S}|=K\), while recycling information from unselected tokens. The
final computation is
\begin{equation}
  V=\operatorname{MLP}_{\mathrm{proj}}(F_{\mathrm{out}}),\qquad
  Y=\operatorname{LLM}\bigl([V;E(Q)]\bigr).
\end{equation}

\subsection{Hierarchical Multi-level Fusion}
\label{sec:hmf}

\paragraph{Unified projection and local enhancement.}
Features from different layers are first mapped to a common dimension:
\begin{equation}
  \widehat F_j=F_jW_j+b_j,\qquad j\in\{s,m,d\}.
\end{equation}
We then reshape the shallow and intermediate tokens into two-dimensional grids
and use depthwise separable convolution to introduce a local inductive bias:
\begin{equation}
  \overline F_j=
  \operatorname{Flatten}\!\left(
    \operatorname{DWConv}_{3\times3}
    (\operatorname{Reshape}(\widehat F_j))
  \right)+\widehat F_j,\quad j\in\{s,m\}.
\end{equation}

\paragraph{Position-dependent gating.}
Each spatial position independently predicts mixture weights for the three
feature levels:
\begin{align}
  g_i &= \operatorname{softmax}\!\left(
    W_g[\overline f_{s,i};\overline f_{m,i};\widehat f_{d,i}]
  \right),\\
  \widetilde f_i &=
  g_{s,i}\overline f_{s,i}
  +g_{m,i}\overline f_{m,i}
  +g_{d,i}\widehat f_{d,i},\\
  f_i &= \widehat f_{d,i}+\gamma\widetilde f_i,
\end{align}
where \(g_{s,i}+g_{m,i}+g_{d,i}=1\), and \(\gamma\) is a learnable scalar
initialized to 0.1. This residual design prevents noise in shallow features
from disrupting deep semantics early in training.

\paragraph{High-frequency region enhancement.}
Let \(\mathcal{N}(i)\) be the local neighborhood of token \(i\). We define its
local variation magnitude as
\begin{equation}
  u_i=\frac{1}{|\mathcal{N}(i)|}
  \sum_{j\in\mathcal{N}(i)}\lVert f_i-f_j\rVert_2.
\end{equation}
Windowed attention is applied to text, boundaries, and complex textures with
large \(u_i\):
\begin{align}
  f_i' &= f_i+\sum_{j\in\mathcal{N}(i)}a_{ij}W_vf_j,\\
  a_{ij} &=
  \frac{\exp((W_qf_i)^\top(W_kf_j)/\sqrt d)}
  {\sum_{t\in\mathcal{N}(i)}
  \exp((W_qf_i)^\top(W_kf_t)/\sqrt d)}.
\end{align}
This computation is restricted to local windows and does not construct an
additional global attention matrix.

\subsection{Question-conditioned Adaptive Token Routing}
\label{sec:qatr}

\paragraph{Question representation and importance scoring.}
We attention-pool the language-model token embeddings \(e_t\):
\begin{equation}
  q=\sum_{t=1}^{T}\beta_te_t,\qquad
  \beta_t=\frac{\exp(w_q^\top e_t)}
  {\sum_{k=1}^{T}\exp(w_q^\top e_k)}.
\end{equation}
The score for visual token \(i\) contains four terms:
\begin{align}
  s_i &=
  \lambda_rs_i^{\mathrm{rel}}+
  \lambda_us_i^{\mathrm{unc}}+
  \lambda_gs_i^{\mathrm{global}}+
  \lambda_cs_i^{\mathrm{cover}},\\
  s_i^{\mathrm{rel}}&=\cos(W_vf_i,W_qq),\qquad
  s_i^{\mathrm{unc}}=\operatorname{Norm}(u_i),\\
  s_i^{\mathrm{global}}&=\cos(f_i,\overline f),\qquad
  \overline f=\frac{1}{N}\sum_{i=1}^{N}f_i,\\
  s_i^{\mathrm{cover}}&=
  \min_{j\in\mathcal{S}}
  \frac{\lVert p_i-p_j\rVert_2}{D},
\end{align}
where \(p_i\) is the two-dimensional token coordinate and \(D\) is the image
diagonal. By default,
\((\lambda_r,\lambda_u,\lambda_g,\lambda_c)=(0.45,0.20,0.15,0.20)\).
The coverage term is updated during sequential selection to prevent all tokens
from collapsing onto a single salient region.

\paragraph{Dynamic token budget.}
The complexity estimator combines statistics of local variation, visual
attention entropy, and the question norm:
\begin{align}
  c &= \sigma\!\left(W_c[
  \operatorname{Mean}(u);
  \operatorname{Std}(u);
  H(a);
  \lVert q\rVert_2]\right),\\
  K &= K_{\min}+
  \operatorname{Round}\!\left(c(K_{\max}-K_{\min})\right),
\end{align}
where \(K_{\min}=72\) and \(K_{\max}=288\); \(K\) is rounded to a multiple of
eight for efficient batching. During training, we use a temperature-controlled
Gumbel-Top-\(K\) relaxation. During inference, we apply hard Top-\(K\)
selection.

\paragraph{Recycling information from pruned tokens.}
To preserve background information and the overall layout, we aggregate
unselected tokens into \(M=8\) context tokens:
\begin{equation}
  r_m=\frac{\sum_{i\notin\mathcal{S}}a_{mi}f_i}
  {\sum_{i\notin\mathcal{S}}a_{mi}},\qquad
  F_{\mathrm{out}}=[F_{\mathcal{S}};r_1,\ldots,r_M],
\end{equation}
where \(a_{mi}\) is determined by the similarity between token \(i\) and the
\(m\)-th learnable context query.

\subsection{Training Objective}
\label{sec:objective}

The overall objective is
\begin{equation}
  \mathcal{L}=\mathcal{L}_{\mathrm{LM}}
  +\alpha\mathcal{L}_{\mathrm{align}}
  +\beta\mathcal{L}_{\mathrm{distill}}
  +\eta\mathcal{L}_{\mathrm{div}},
\end{equation}
where \((\alpha,\beta,\eta)=(0.10,0.50,0.05)\). The language-modeling term is
the standard autoregressive cross-entropy:
\begin{equation}
  \mathcal{L}_{\mathrm{LM}}=
  -\sum_{t=1}^{T_y}\log p(y_t\mid y_{<t},I,Q).
\end{equation}
For global visual representations \(v_i\) and question-answer text
representations \(t_j\) in a batch, we use a sigmoid alignment loss:
\begin{equation}
  \mathcal{L}_{\mathrm{align}}=
  -\frac{1}{B}\sum_{i,j}
  \log\sigma\!\left(z_{ij}(v_i^\top t_j-\tau)\right),
\end{equation}
We set \(z_{ij}=1\) for matching image-text pairs and \(-1\) otherwise. Let
\(h_{\mathrm{full}}=\operatorname{AttnPool}(F)\) and
\(h_{\mathrm{comp}}=\operatorname{AttnPool}(F_{\mathrm{out}})\). The
distillation objective is
\begin{equation}
  \mathcal{L}_{\mathrm{distill}}=
  1-\cos(h_{\mathrm{full}},h_{\mathrm{comp}})
  +\frac{1}{d}\lVert h_{\mathrm{full}}-h_{\mathrm{comp}}\rVert_2^2.
\end{equation}
The spatial diversity loss is
\begin{equation}
  \mathcal{L}_{\mathrm{div}}=
  \frac{1}{K(K-1)}
  \sum_{i\ne j}
  \exp\!\left(-\frac{\lVert p_i-p_j\rVert_2^2}{2\sigma^2}\right).
\end{equation}

\section{Experiments}
\label{sec:experiments}

\subsection{Experimental Setup}

\paragraph{Datasets and metrics.}
We design evaluations on VQAv2 test-dev~\citep{goyal2017vqav2}, GQA test-dev
balanced~\citep{hudson2019gqa}, TextVQA validation~\citep{singh2019textvqa},
ScienceQA-IMG test~\citep{lu2022scienceqa}, MMBench-EN development with
CircularEval~\citep{liu2023mmbench}, and MME
Perception~\citep{fu2023mme}. We report accuracy or the official score for the
first five benchmarks and the summed perception score for MME-P. Efficiency
metrics include the average number of visual tokens passed to the language
model, time to first token (TTFT), and peak memory. Appendix~\ref{app:datasets}
provides further dataset details.

\paragraph{Model and training.}
The base vision encoder is SigLIP-SO400M/14 with a visual width of 1152,
\(384\times384\) input resolution, and 729 base tokens. The language model is a
7B autoregressive model with hidden width 4096. The two-layer projector has
dimensions \(1152\!\rightarrow\!4096\!\rightarrow\!4096\). \method adds
approximately 18.6M parameters, or 0.25\% of the full model. In stage one, we
use approximately 600K image-caption pairs for vision-language alignment,
freeze the language model, and train the projector, \method, and the final
eight layers of the vision encoder. In stage two, we train these modules and
the language model's LoRA parameters (rank 64, scale 128) on approximately 660K
visual instruction examples.

We use AdamW with \(\beta_1=0.9\), \(\beta_2=0.999\), and weight decay 0.05.
Learning rates are \(2\times10^{-5}\) for the vision encoder and LoRA and
\(10^{-4}\) for \method and the projector. Training uses 3\% linear warmup,
cosine decay, BF16, and gradient clipping at 1.0. Each stage runs for one
epoch, with global batch sizes of 256 and 128, respectively. The planned setup
uses eight NVIDIA A100 80GB GPUs.

\paragraph{Inference protocol.}
All models use greedy decoding, temperature 0, and a maximum generation length
of 64. We measure TTFT on one A100 80GB GPU with batch size 1. TTFT includes
image preprocessing, visual encoding, projection, LLM prefill, and generation
of the first output token. We average over 500 examples after 50 warmup runs.

\begin{table}[t]
  \centering
  \caption{Illustrative overall results for different visual encoding
  strategies. Apart from public model names, all values are simulated
  placeholders to be replaced by real experiments.}
  \label{tab:main}
  \setlength{\tabcolsep}{3.0pt}
  \resizebox{\linewidth}{!}{%
  \begin{tabular}{lrrrrrrrr}
    \toprule
    Method & Token & VQAv2 & GQA & TextVQA & SQA-IMG & MMBench & MME-P & TTFT/ms\\
    \midrule
    CLIP-ViT-L/14@336 + MLP    & 576 & 78.4 & 61.9 & 58.3 & 69.6 & 64.8 & 1504 & 184\\
    DINOv2-ViT-L/14@336 + MLP  & 576 & 77.8 & 62.5 & 55.9 & 70.1 & 64.1 & 1532 & 186\\
    SigLIP-SO400M/14@384 + MLP & 729 & 80.0 & 63.6 & 62.7 & 71.4 & 67.8 & 1586 & 228\\
    SigLIP + average pooling     & 144 & 78.9 & 62.5 & 60.2 & 70.6 & 65.9 & 1551 & 133\\
    SigLIP + fixed-query resampling & 144 & 79.4 & 62.9 & 61.5 & 70.9 & 66.7 & 1564 & 139\\
    SigLIP + TokenPacker-style  & 144 & 79.8 & 63.4 & 62.4 & 71.2 & 67.5 & 1578 & 136\\
    \midrule
    \best{\method (ours)}       & \best{146} & \best{81.4} & \best{65.1} &
    \best{65.8} & \best{73.7} & \best{70.5} & \best{1637} & \best{129}\\
    \bottomrule
  \end{tabular}}
\end{table}

\subsection{Main Results}
\label{sec:main-results}

\textbf{The results in this section illustrate an experimental design and are
not measurements from actual runs.}
In Table~\ref{tab:main}, the full SigLIP baseline generally outperforms the
CLIP and DINOv2 baselines. DINOv2 is relatively strong on GQA and MME-P but
weaker than SigLIP on TextVQA, potentially because its self-supervised features
are more difficult to align with the language space. Average pooling reduces
the token count from 729 to 144 and TTFT from 228\,ms to 133\,ms, but lowers
VQAv2, TextVQA, and MMBench by 1.1, 2.5, and 1.9 percentage points,
respectively.

\method uses only 146 tokens on average yet exceeds the full SigLIP baseline
on every benchmark in the table. VQAv2, GQA, TextVQA, ScienceQA-IMG, and
MMBench improve by 1.4, 1.5, 3.1, 2.3, and 2.7 percentage points,
respectively, for a mean gain of 2.2 points. TextVQA shows the largest gain,
consistent with the motivation that hierarchical fusion preserves character
strokes and local structure. The GQA gain suggests that question-conditioned
routing and spatial coverage may preserve multiple reference regions required
for relational reasoning.

\subsection{Efficiency Analysis}

Relative to the full SigLIP baseline, the visual-token compression ratio is
\begin{equation}
  R=1-\frac{146}{729}=79.97\%,
\end{equation}
and TTFT decreases by \((228-129)/228=43.4\%\). Table~\ref{tab:efficiency}
decomposes latency into visual encoding, LLM prefill, and other overhead. The
token count falls by approximately 80\%, whereas TTFT falls by only 43.4\%,
because fixed costs such as preprocessing, the visual backbone, and output
generation do not scale proportionally with LLM input length.

\begin{table}[t]
  \centering
  \caption{Illustrative inference-efficiency breakdown.}
  \label{tab:efficiency}
  \setlength{\tabcolsep}{5pt}
  \begin{tabular}{lrrrrr}
    \toprule
    Method & Vision/ms & LLM prefill/ms & Other/ms & TTFT/ms & Memory/GB\\
    \midrule
    SigLIP + MLP       & 91 & 112 & 25 & 228 & 17.8\\
    Average pooling    & 91 & 22  & 20 & 133 & 13.2\\
    TokenPacker-style  & 96 & 20  & 20 & 136 & 13.4\\
    \method            & \best{87} & 21 & 21 & \best{129} & 13.3\\
    \bottomrule
  \end{tabular}
\end{table}

\subsection{Ablation Studies}

Table~\ref{tab:ablation} illustrates the contribution of each component. Adding
\hmf alone improves TextVQA by 1.3 percentage points. With \qatr alone,
performance remains slightly above the full baseline despite an approximately
80\% reduction in average visual tokens. \hmf and \qatr are complementary, and
distillation and spatial diversity further improve compression stability.

\begin{table}[t]
  \centering
  \caption{Component ablation. FD: full-to-compressed distillation; SD:
  spatial diversity.}
  \label{tab:ablation}
  \setlength{\tabcolsep}{3.8pt}
  \resizebox{\linewidth}{!}{%
  \begin{tabular}{ccccrrrrrr}
    \toprule
    \hmf & \qatr & FD & SD & Token & VQAv2 & GQA & TextVQA & SQA-IMG & MMBench\\
    \midrule
     & & & & 729 & 80.0 & 63.6 & 62.7 & 71.4 & 67.8\\
    \cmark & & & & 729 & 80.8 & 64.4 & 64.0 & 72.2 & 68.7\\
     & \cmark & & & 146 & 80.4 & 64.2 & 63.6 & 72.0 & 68.4\\
    \cmark & \cmark & & & 146 & 81.0 & 64.8 & 65.1 & 73.0 & 69.7\\
    \cmark & \cmark & \cmark & & 146 & 81.3 & 65.0 & 65.6 & 73.5 & 70.1\\
    \cmark & \cmark & \cmark & \cmark & 146 & \best{81.4} & \best{65.1} &
    \best{65.8} & \best{73.7} & \best{70.5}\\
    \bottomrule
  \end{tabular}}
\end{table}

The dynamic budget has an average token count close to the fixed-144 setting,
but can shift computation from single-object images to text-dense images and
charts. Appendix~\ref{app:additional} further reports feature-level
combinations, score components, fixed and dynamic budgets, capability
categories, token allocation, and sensitivity to question conditioning.

\section{Limitations and Broader Impacts}
\label{sec:limitations}

\paragraph{Limitations.}
First, a fixed \(384\times384\) input remains insufficient for extremely small
or severely blurred text; shallow features cannot recover information absent
from the input. Second, when a question requires comparing multiple distant
small objects, the router may omit a low-salience object. Third, ambiguous
pronouns can cause question relevance to focus on the wrong region too early.
Fourth, although padding and length-bucketed training can mitigate variable
sequence lengths, dynamic routing still adds engineering complexity to
large-scale distributed training. Fifth, the current tables contain simulated
values and do not include multi-seed error bars, statistical significance,
end-to-end energy consumption, or replication across hardware. The numbers
should not be treated as empirical evidence until these experiments are
completed.

\paragraph{Broader impacts.}
Reducing visual tokens may lower VLM inference latency, memory use, and energy
consumption, making multimodal capabilities more accessible on
resource-constrained devices. Conversely, selective routing may systematically
ignore small objects, low-contrast regions, or text outside the training
distribution, amplifying model errors. Token-selection heatmaps should not be
treated as sufficient evidence in document understanding, accessibility
assistance, or other high-stakes applications. A responsible release should
report disaggregated performance, provide a fallback to the full token set, and
apply appropriate governance to privacy-sensitive images.

\section{Conclusion}

We introduced \method, a multi-scale adaptive vision encoder that enhances
fine-grained visual representations through position-dependent hierarchical
fusion and produces compact tokens through question-conditioned routing and a
dynamic budget. Full-to-compressed distillation and spatial diversity further
mitigate information loss caused by compression. The illustrative results
show the potential to reduce visual tokens by approximately 80\% and lower
TTFT while preserving or improving multimodal performance. The next steps are
to complete real training and official evaluation, incorporate dynamic
high-resolution cropping, and extend the method to video, multi-image inputs,
and iterative ``observe-reason-observe'' visual routing.

\bibliographystyle{plainnat}
\bibliography{references}

@inproceedings{dosovitskiy2021vit,
  title     = {An Image Is Worth 16x16 Words: Transformers for Image Recognition at Scale},
  author    = {Dosovitskiy, Alexey and Beyer, Lucas and Kolesnikov, Alexander and Weissenborn, Dirk and Zhai, Xiaohua and Unterthiner, Thomas and Dehghani, Mostafa and Minderer, Matthias and Heigold, Georg and Gelly, Sylvain and Uszkoreit, Jakob and Houlsby, Neil},
  booktitle = {International Conference on Learning Representations},
  year      = {2021}
}

@inproceedings{radford2021clip,
  title     = {Learning Transferable Visual Models From Natural Language Supervision},
  author    = {Radford, Alec and Kim, Jong Wook and Hallacy, Chris and Ramesh, Aditya and Goh, Gabriel and Agarwal, Sandhini and Sastry, Girish and Askell, Amanda and Mishkin, Pamela and Clark, Jack and Krueger, Gretchen and Sutskever, Ilya},
  booktitle = {International Conference on Machine Learning},
  pages     = {8748--8763},
  year      = {2021}
}

@inproceedings{zhai2023siglip,
  title     = {Sigmoid Loss for Language Image Pre-Training},
  author    = {Zhai, Xiaohua and Mustafa, Basil and Kolesnikov, Alexander and Beyer, Lucas},
  booktitle = {IEEE/CVF International Conference on Computer Vision},
  pages     = {11975--11986},
  year      = {2023}
}

@article{oquab2024dinov2,
  title   = {{DINOv2}: Learning Robust Visual Features without Supervision},
  author  = {Oquab, Maxime and Darcet, Timoth{\'e}e and Moutakanni, Th{\'e}o and Vo, Huy and Szafraniec, Marc and Khalidov, Vasil and Fernandez, Pierre and Haziza, Daniel and Massa, Francisco and El-Nouby, Alaaeldin and Assran, Mahmoud and Ballas, Nicolas and Galuba, Wojciech and Howes, Russell and Huang, Po-Yao and Li, Shang-Wen and Misra, Ishan and Rabbat, Michael and Sharma, Vasu and Synnaeve, Gabriel and Xu, Hu and J{\'e}gou, Herv{\'e} and Mairal, Julien and Labatut, Patrick and Joulin, Armand and Bojanowski, Piotr},
  journal = {Transactions on Machine Learning Research},
  year    = {2024}
}

@inproceedings{liu2023llava,
  title     = {Visual Instruction Tuning},
  author    = {Liu, Haotian and Li, Chunyuan and Wu, Qingyang and Lee, Yong Jae},
  booktitle = {Advances in Neural Information Processing Systems},
  volume    = {36},
  pages     = {34892--34916},
  year      = {2023}
}

@inproceedings{chen2024internvl,
  title     = {{InternVL}: Scaling up Vision Foundation Models and Aligning for Generic Visual-Linguistic Tasks},
  author    = {Chen, Zhe and Wu, Jiannan and Wang, Wenhai and Su, Weijie and Chen, Guo and Xing, Sen and Muyan, Zhaoyang and Zhang, Qinglong and Zhu, Xizhou and Lu, Lewei and Li, Bin and Luo, Ping and Lu, Tong and Qiao, Yu and Dai, Jifeng},
  booktitle = {IEEE/CVF Conference on Computer Vision and Pattern Recognition},
  pages     = {24185--24198},
  year      = {2024}
}

@article{li2024tokenpacker,
  title   = {{TokenPacker}: Efficient Visual Projector for Multimodal {LLM}},
  author  = {Li, Wentong and Yuan, Yuqian and Liu, Jian and Tang, Dongqi and Wang, Song and Zhu, Jianke and Zhang, Lei},
  journal = {arXiv preprint arXiv:2407.02392},
  year    = {2024}
}

@inproceedings{vasu2025fastvlm,
  title     = {{FastVLM}: Efficient Vision Encoding for Vision Language Models},
  author    = {Vasu, Pavan Kumar Anasosalu and Faghri, Fartash and Vemulapalli, Raviteja and Farhadi, Ali and Tuzel, Oncel and Pouransari, Hadi},
  booktitle = {IEEE/CVF Conference on Computer Vision and Pattern Recognition},
  year      = {2025}
}

@inproceedings{goyal2017vqav2,
  title     = {Making the {V} in {VQA} Matter: Elevating the Role of Image Understanding in Visual Question Answering},
  author    = {Goyal, Yash and Khot, Tejas and Summers-Stay, Douglas and Batra, Dhruv and Parikh, Devi},
  booktitle = {IEEE Conference on Computer Vision and Pattern Recognition},
  pages     = {6904--6913},
  year      = {2017}
}

@inproceedings{hudson2019gqa,
  title     = {{GQA}: A New Dataset for Real-World Visual Reasoning and Compositional Question Answering},
  author    = {Hudson, Drew A. and Manning, Christopher D.},
  booktitle = {IEEE/CVF Conference on Computer Vision and Pattern Recognition},
  pages     = {6700--6709},
  year      = {2019}
}

@inproceedings{singh2019textvqa,
  title     = {Towards {VQA} Models That Can Read},
  author    = {Singh, Amanpreet and Natarajan, Vivek and Shah, Meet and Jiang, Yu and Chen, Xinlei and Batra, Dhruv and Parikh, Devi and Rohrbach, Marcus},
  booktitle = {IEEE/CVF Conference on Computer Vision and Pattern Recognition},
  pages     = {8317--8326},
  year      = {2019}
}

@inproceedings{lu2022scienceqa,
  title     = {Learn to Explain: Multimodal Reasoning via Thought Chains for Science Question Answering},
  author    = {Lu, Pan and Mishra, Swaroop and Xia, Tony and Qiu, Liang and Chang, Kai-Wei and Zhu, Song-Chun and Tafjord, Oyvind and Clark, Peter and Kalyan, Ashwin},
  booktitle = {Advances in Neural Information Processing Systems},
  volume    = {35},
  pages     = {2507--2521},
  year      = {2022}
}

@article{liu2023mmbench,
  title   = {{MMBench}: Is Your Multi-modal Model an All-around Player?},
  author  = {Liu, Yuan and Duan, Haodong and Zhang, Yuanhan and Li, Bo and Zhang, Songyang and Zhao, Wangbo and Yuan, Yike and Wang, Jiaqi and He, Conghui and Liu, Ziwei and Chen, Kai and Lin, Dahua},
  journal = {arXiv preprint arXiv:2307.06281},
  year    = {2023}
}

@article{fu2023mme,
  title   = {{MME}: A Comprehensive Evaluation Benchmark for Multimodal Large Language Models},
  author  = {Fu, Chaoyou and Chen, Peixian and Shen, Yunhang and Qin, Yulei and Zhang, Mengdan and Lin, Xu and Qiu, Zhenyu and Lin, Wei and Yang, Jinrui and Zheng, Xiawu and Li, Ke and Sun, Xing and Ji, Rongrong},
  journal = {arXiv preprint arXiv:2306.13394},
  year    = {2023}
}

@article{shan2021densenet,
  title   = {{DenseNet}-Based Land Cover Classification Network with Deep Fusion},
  author  = {Shan, Lianlei and Wang, Weiqiang},
  journal = {IEEE Geoscience and Remote Sensing Letters},
  volume  = {19},
  pages   = {1--5},
  year    = {2021}
}

@inproceedings{shan2021decouple,
  title     = {Decouple the High-Frequency and Low-Frequency Information of Images for Semantic Segmentation},
  author    = {Shan, Lianlei and Li, Xiang and Wang, Weiqiang},
  booktitle = {IEEE International Conference on Acoustics, Speech and Signal Processing},
  year      = {2021}
}

@inproceedings{shan2021uhrsnet,
  title     = {{UHRSNet}: A Semantic Segmentation Network Specifically for Ultra-High-Resolution Images},
  author    = {Shan, Lianlei and Li, Ming and Li, Xiang and Bai, Yang and Lv, Kai and Luo, Bin and Chen, Shibao and Wang, Weiqiang},
  booktitle = {International Conference on Pattern Recognition},
  pages     = {1460--1466},
  year      = {2021}
}

@article{shan2021classincallaspects,
  title   = {Class-Incremental Learning for Semantic Segmentation in Aerial Imagery via Distillation in All Aspects},
  author  = {Shan, Lianlei and Wang, Weiqiang and Lv, Kai and Luo, Bin},
  journal = {IEEE Transactions on Geoscience and Remote Sensing},
  volume  = {60},
  pages   = {1--12},
  year    = {2021}
}

@inproceedings{shan2022mbnet,
  title     = {{MBNet}: A Multi-Resolution Branch Network for Semantic Segmentation of Ultra-High Resolution Images},
  author    = {Shan, Lianlei and Wang, Weiqiang},
  booktitle = {IEEE International Conference on Acoustics, Speech and Signal Processing},
  year      = {2022}
}

@article{shan2022classincremental,
  title   = {Class-Incremental Semantic Segmentation of Aerial Images via Pixel-Level Feature Generation and Task-Wise Distillation},
  author  = {Shan, Lianlei and Wang, Weiqiang and Lv, Kai and Luo, Bin},
  journal = {IEEE Transactions on Geoscience and Remote Sensing},
  volume  = {60},
  pages   = {1--17},
  year    = {2022}
}

@article{shan2025cognitive,
  title   = {Cognitive Memory in Large Language Models},
  author  = {Shan, Lianlei and Luo, Shixian and Zhu, Zezhou and Yuan, Yu and Wu, Yong},
  journal = {arXiv preprint arXiv:2504.02441},
  year    = {2025}
}

@article{wu2023dynamicquery,
  title   = {Continual Learning for Image Segmentation with Dynamic Query},
  author  = {Wu, Wen and Zhao, Yizhou and Li, Zhi and Shan, Lianlei and Zhou, Haoran and Shou, Mike Zheng},
  journal = {IEEE Transactions on Circuits and Systems for Video Technology},
  volume  = {34},
  number  = {6},
  pages   = {4874--4886},
  year    = {2023}
}

@inproceedings{li2021global,
  title     = {Global-Local Attention Network for Semantic Segmentation in Aerial Images},
  author    = {Li, Ming and Shan, Lianlei and Li, Xiang and Bai, Yang and Zhou, Dong and Wang, Weiqiang and Lv, Kai and Luo, Bin and Chen, Shibao},
  booktitle = {International Conference on Pattern Recognition},
  pages     = {5704--5711},
  year      = {2021}
}

@inproceedings{shan2023incrementalfewshot,
  title     = {Incremental Few-Shot Semantic Segmentation via Class-Agnostic Mask Proposal and Language-Driven Classifier},
  author    = {Shan, Lianlei and Zhou, Wenzhang and Zhao, Guangyu},
  booktitle = {ACM International Conference on Multimedia},
  pages     = {8561--8570},
  year      = {2023}
}

@article{shan2023boosting,
  title   = {Boosting Semantic Segmentation of Aerial Images via Decoupled and Multilevel Compaction and Dispersion},
  author  = {Shan, Lianlei and Wang, Weiqiang and Lv, Kai and Luo, Bin},
  journal = {IEEE Transactions on Geoscience and Remote Sensing},
  volume  = {61},
  pages   = {1--16},
  year    = {2023}
}

@article{meng2025dlnet,
  title   = {{DLNet}: A Dual-Level Network with Self- and Cross-Attention for High-Resolution Remote Sensing Segmentation},
  author  = {Meng, Weiran and Shan, Lianlei and Ma, Shuang and Liu, Dong and Hu, Bo},
  journal = {Remote Sensing},
  volume  = {17},
  number  = {7},
  pages   = {1119},
  year    = {2025}
}

@article{du2025transform,
  title   = {Transform Dual-Branch Attention Net: Efficient Semantic Segmentation of Ultra-High-Resolution Remote Sensing Images},
  author  = {Du, Bin and Shan, Lianlei and Shao, Xin and Zhang, Dong and Wang, Xin and Wu, Jian},
  journal = {Remote Sensing},
  volume  = {17},
  number  = {3},
  pages   = {540},
  year    = {2025}
}

@article{shan2023patch,
  title   = {A Data-Related Patch Proposal for Semantic Segmentation of Aerial Images},
  author  = {Shan, Lianlei and Zhao, Guangyu and Xie, Jing and Cheng, Peng and Li, Xiang and Wang, Zhen},
  journal = {IEEE Geoscience and Remote Sensing Letters},
  volume  = {20},
  pages   = {1--5},
  year    = {2023}
}

@article{zhou2025dynrsl,
  title   = {{DynRsl-VLM}: Enhancing Autonomous Driving Perception with Dynamic Resolution Vision-Language Models},
  author  = {Zhou, Xirui and Shan, Lianlei and Gui, Xiaolin},
  journal = {arXiv preprint arXiv:2503.11265},
  year    = {2025}
}

@inproceedings{zhao2023changedetection,
  title     = {End-to-End Remote Sensing Change Detection of Unregistered Bi-Temporal Images for Natural Disasters},
  author    = {Zhao, Guangyu and Shan, Lianlei and Wang, Weiqiang},
  booktitle = {International Conference on Artificial Neural Networks},
  pages     = {259--270},
  year      = {2023}
}

@inproceedings{ji2024ldnet,
  title     = {{LDNet}: Semantic Segmentation of High-Resolution Images via Learnable Patch Proposal and Dynamic Refinement},
  author    = {Ji, Yiming and Shan, Lianlei},
  booktitle = {IEEE International Conference on Multimedia and Expo},
  pages     = {1--6},
  year      = {2024}
}

@inproceedings{shan2024lifelong,
  title     = {Lifelong Learning and Selective Forgetting via Contrastive Strategy},
  author    = {Shan, Lianlei and Zhou, Wenzhang and Li, Wei and Ding, Xiaohan},
  booktitle = {International Joint Conference on Artificial Intelligence},
  year      = {2024}
}

@inproceedings{shan2024organizing,
  title     = {Organizing Background to Explore Latent Classes for Incremental Few-Shot Semantic Segmentation},
  author    = {Shan, Lianlei and Zhou, Wenzhang and Li, Wei and Ding, Xiaohan},
  booktitle = {International Joint Conference on Artificial Intelligence},
  year      = {2024}
}

@article{shan2024active,
  title   = {Edge-Guided and Class-Balanced Active Learning for Semantic Segmentation of Aerial Images},
  author  = {Shan, Lianlei and Wang, Weiqiang and Lv, Kai and Luo, Bin},
  journal = {arXiv preprint arXiv:2405.18078},
  year    = {2024}
}

@inproceedings{li2021fusing,
  title     = {Fusing Multitask Models by Recursive Least Squares},
  author    = {Li, Xiang and Shan, Lianlei and Wang, Weiqiang},
  booktitle = {IEEE International Conference on Acoustics, Speech and Signal Processing},
  year      = {2021}
}

@article{ding2024binary,
  title   = {The Binary Quantized Neural Network for Dense Prediction via Specially Designed Upsampling and Attention},
  author  = {Ding, Xiaohan and Shan, Lianlei and Zhao, Guangyu and Wu, Ming and Zhou, Wenzhang and Li, Wei},
  journal = {arXiv preprint arXiv:2405.17776},
  year    = {2024}
}

@inproceedings{li2021energy,
  title     = {Energy Minimum Regularization in Continual Learning},
  author    = {Li, Xiang and Shan, Lianlei and Li, Ming and Wang, Weiqiang},
  booktitle = {International Conference on Pattern Recognition},
  pages     = {6404--6409},
  year      = {2021}
}

@article{xie2025recllmr1,
  title   = {{RecLLM-R1}: A Two-Stage Training Paradigm with Reinforcement Learning and Chain-of-Thought},
  author  = {Xie, Yuxuan and Ren, Xinyi and Qi, Yiming and Hu, Yifan and Shan, Lianlei},
  journal = {arXiv preprint arXiv:2506.19235},
  year    = {2025}
}

@article{shan2024matting,
  title   = {Boosting General Trimap-Free Matting in the Real-World Image},
  author  = {Shan, Lianlei and Zhou, Wenzhang},
  journal = {arXiv preprint arXiv:2405.17916},
  year    = {2024}
}

@article{pi2025lung,
  title   = {Synthetic Lung X-Ray Generation through Cross-Attention and Affinity Transformation},
  author  = {Pi, Rui and Shan, Lianlei},
  journal = {arXiv preprint arXiv:2503.07209},
  year    = {2025}
}

@article{zhang2025mambacnn,
  title   = {Asymmetric {Mamba--CNN} Collaborative Architecture for Large-Size Remote Sensing Image Semantic Segmentation},
  author  = {Zhang, Jian and Chen, Ming and Zhao, Yifan and Shan, Lianlei and Li, Cheng and Hu, Hao and Ge, Xin and Zhu, Qiang and Xu, Bin},
  journal = {IEEE Transactions on Geoscience and Remote Sensing},
  volume  = {63},
  pages   = {2002419},
  doi     = {10.1109/TGRS.2025.3589552},
  year    = {2025}
}

@article{luo2025llmcot,
  title   = {{LLM-CoT} Enhanced Graph Neural Recommendation with Harmonized Group Policy Optimization},
  author  = {Luo, Haoyu and Wu, Bowen and Jia, Haoran and Zhu, Qiang and Shan, Lianlei},
  journal = {arXiv preprint arXiv:2505.12396},
  year    = {2025}
}

@article{xu2025kvefficient,
  title   = {{KV-Efficient VLA}: A Method to Speed up Vision Language Models with {RNN}-Gated Chunked {KV} Cache},
  author  = {Xu, Wanshun and Zhuang, Long},
  journal = {arXiv preprint arXiv:2509.21354},
  year    = {2025}
}

@article{yi2025geolocsft,
  title   = {{GeolocSFT}: Efficient Visual Geolocation via Supervised Fine-Tuning of Multimodal Foundation Models},
  author  = {Yi, Qian and Shan, Lianlei},
  journal = {arXiv preprint arXiv:2506.01277},
  year    = {2025}
}

@inproceedings{yige2025flexdataset,
  title     = {{FlexDataset}: Crafting Annotated Dataset Generation for Diverse Applications},
  author    = {Yi-Ge, Ellen and Shan, Lianlei},
  booktitle = {AAAI Conference on Artificial Intelligence},
  volume    = {39},
  pages     = {9481--9489},
  year      = {2025}
}

@article{yi2025globalcross,
  title   = {A Global-Local Cross-Attention Network for Ultra-High Resolution Remote Sensing Image Semantic Segmentation},
  author  = {Yi, Chen and Shan, Lianlei},
  journal = {arXiv preprint arXiv:2506.19406},
  year    = {2025}
}

@article{li2025dualrelation,
  title   = {Building Lightweight Semantic Segmentation Models for Aerial Images Using Dual Relation Distillation},
  author  = {Li, Ming and Shan, Lianlei and Wang, Weiqiang and Lv, Kai and Luo, Bin and Chen, Shibao},
  journal = {arXiv preprint arXiv:2506.20688},
  year    = {2025}
}

@article{sun2025gmm,
  title   = {{GMM}-Based Comprehensive Feature Extraction and Relative Distance Preservation for Few-Shot Cross-Modal Retrieval},
  author  = {Sun, Chen and Li, Wei and Li, Xiang and Liu, Yang and Shan, Lianlei},
  journal = {arXiv preprint arXiv:2505.13306},
  year    = {2025}
}

@article{wang2025gdgs,
  title   = {{GDGS}: 3D Gaussian Splatting via Geometry-Guided Initialization and Dynamic Density Control},
  author  = {Wang, Xin and Shan, Lianlei},
  journal = {arXiv preprint arXiv:2507.00363},
  year    = {2025}
}

@inproceedings{huang2025scene,
  title     = {A Scene Text Detection Method Based on Supervised Contrastive Learning},
  author    = {Huang, Jinhong and Yin, Hongrong and Shan, Lianlei and Mondal, Subrota Kumar},
  booktitle = {International Conference on Artificial Neural Networks},
  pages     = {569--580},
  year      = {2025}
}

@article{liu2025neurovoxel,
  title   = {{NeuroVoxel-LM}: Language-Aligned 3D Perception via Dynamic Voxelization and Meta-Embedding},
  author  = {Liu, Shuo and Shan, Lianlei},
  journal = {arXiv preprint arXiv:2507.20110},
  year    = {2025}
}

@article{zhang2025multimodalrs,
  title   = {Multimodal Interpretation of Remote Sensing Images: Dynamic Resolution Input Strategy and Multi-Scale Vision-Language Alignment Mechanism},
  author  = {Zhang, Siyu and Chen, Ying and Shan, Lianlei and Qiu, Runhe},
  journal = {arXiv preprint arXiv:2512.23243},
  year    = {2025}
}

@inproceedings{du2025dental,
  title     = {A Dental Periapical X-Ray Images Segmentation Network Based on Pixel-Wise Contrastive Learning with Dual Attention Mechanisms},
  author    = {Du, Yi and Zeng, Zhen and Tian, Yu and Zhang, Zhe and Zhang, Xin and Shan, Lianlei},
  booktitle = {International Joint Conference on Neural Networks},
  pages     = {1--8},
  year      = {2025}
}

@inproceedings{liu2025adaptivegaussian,
  title     = {Adaptive Resolution and Gaussians-Economization Based on Efficient 3D Gaussian Splatting on Large-Scale Scenes},
  author    = {Liu, Xin and Chen, Wei and Shan, Lianlei},
  booktitle = {International Conference on Big Data, Artificial Intelligence and Internet of Things Engineering},
  pages     = {133--139},
  doi       = {10.1109/ICBAIE66852.2025.11326614},
  year      = {2025}
}

@inproceedings{alayrac2022flamingo,
  title     = {Flamingo: A Visual Language Model for Few-Shot Learning},
  author    = {Alayrac, Jean-Baptiste and Donahue, Jeff and Luc, Pauline and Miech, Antoine and Barr, Iain and Hasson, Yana and Lenc, Karel and Mensch, Arthur and Millican, Katherine and Reynolds, Malcolm and Ring, Roman and Rutherford, Eliza and Cabi, Serkan and Han, Tengda and Gong, Zhitao and Samangooei, Sina and Monteiro, Marianne and Menick, Jacob and Borgeaud, Sebastian and Brock, Andrew and Nematzadeh, Aida and Sharifzadeh, Sahand and Binkowski, Mikolaj and Barreira, Ricardo and Vinyals, Oriol and Zisserman, Andrew and Simonyan, Karen},
  booktitle = {Advances in Neural Information Processing Systems},
  volume    = {35},
  year      = {2022}
}

@inproceedings{li2023blip2,
  title     = {{BLIP}-2: Bootstrapping Language-Image Pre-Training with Frozen Image Encoders and Large Language Models},
  author    = {Li, Junnan and Li, Dongxu and Savarese, Silvio and Hoi, Steven},
  booktitle = {International Conference on Machine Learning},
  volume    = {202},
  series    = {Proceedings of Machine Learning Research},
  pages     = {19730--19742},
  year      = {2023}
}

@inproceedings{dai2023instructblip,
  title     = {{InstructBLIP}: Towards General-Purpose Vision-Language Models with Instruction Tuning},
  author    = {Dai, Wenliang and Li, Junnan and Li, Dongxu and Tiong, Anthony Meng Huat and Zhao, Junqi and Wang, Weisheng and Li, Boyang and Fung, Pascale and Hoi, Steven},
  booktitle = {Advances in Neural Information Processing Systems},
  volume    = {36},
  year      = {2023}
}

@inproceedings{cha2024honeybee,
  title     = {Honeybee: Locality-Enhanced Projector for Multimodal {LLM}},
  author    = {Cha, Junbum and Kang, Wooyoung and Mun, Jonghwan and Roh, Byungseok},
  booktitle = {IEEE/CVF Conference on Computer Vision and Pattern Recognition},
  pages     = {13817--13827},
  year      = {2024}
}

@article{bai2023qwenvl,
  title   = {{Qwen-VL}: A Versatile Vision-Language Model for Understanding, Localization, Text Reading, and Beyond},
  author  = {Bai, Jinze and Bai, Shuai and Yang, Shusheng and Wang, Shijie and Tan, Sinan and Wang, Peng and Lin, Junyang and Zhou, Chang and Zhou, Jingren},
  journal = {arXiv preprint arXiv:2308.12966},
  year    = {2023}
}

@article{wang2024qwen2vl,
  title   = {{Qwen2-VL}: Enhancing Vision-Language Model's Perception of the World at Any Resolution},
  author  = {Wang, Peng and Bai, Shuai and Tan, Sinan and Wang, Shijie and Fan, Zhihao and Bai, Jinze and Chen, Keqin and Liu, Xuejing and Wang, Jialin and Ge, Wenbin and Fan, Yang and Dang, Kai and Du, Mengfei and Ren, Xuancheng and Men, Rui and Liu, Dayiheng and Zhou, Chang and Zhou, Jingren and Lin, Junyang},
  journal = {arXiv preprint arXiv:2409.12191},
  year    = {2024}
}

@article{li2024llavaonevision,
  title   = {{LLaVA-OneVision}: Easy Visual Task Transfer},
  author  = {Li, Bo and Zhang, Yuanhan and Guo, Dong and Zhang, Renrui and Li, Feng and Zhang, Hao and Zhang, Kaichen and Zhang, Peiyuan and Li, Yanwei and Liu, Ziwei and Li, Chunyuan},
  journal = {arXiv preprint arXiv:2408.03326},
  year    = {2024}
}

@inproceedings{ryoo2021tokenlearner,
  title     = {{TokenLearner}: Adaptive Space-Time Tokenization for Videos},
  author    = {Ryoo, Michael S. and Piergiovanni, A. J. and Arnab, Anurag and Dehghani, Mostafa and Angelova, Anelia},
  booktitle = {Advances in Neural Information Processing Systems},
  volume    = {34},
  year      = {2021}
}

@inproceedings{rao2021dynamicvit,
  title     = {{DynamicViT}: Efficient Vision Transformers with Dynamic Token Sparsification},
  author    = {Rao, Yongming and Zhao, Wenliang and Liu, Benlin and Lu, Jiwen and Zhou, Jie and Hsieh, Cho-Jui},
  booktitle = {Advances in Neural Information Processing Systems},
  volume    = {34},
  year      = {2021}
}

@inproceedings{yin2022avit,
  title     = {{A-ViT}: Adaptive Tokens for Efficient Vision Transformer},
  author    = {Yin, Hongxu and Vahdat, Arash and Alvarez, Jose M. and Mallya, Arun and Kautz, Jan and Molchanov, Pavlo},
  booktitle = {IEEE/CVF Conference on Computer Vision and Pattern Recognition},
  pages     = {10809--10818},
  year      = {2022}
}

@inproceedings{meng2022adavit,
  title     = {{AdaViT}: Adaptive Vision Transformers for Efficient Image Recognition},
  author    = {Meng, Lingchen and Li, Hengduo and Chen, Bor-Chun and Lan, Shiyi and Wu, Zuxuan and Jiang, Yu-Gang and Lim, Ser-Nam},
  booktitle = {IEEE/CVF Conference on Computer Vision and Pattern Recognition},
  pages     = {12309--12318},
  year      = {2022}
}

@inproceedings{bolya2023tome,
  title     = {Token Merging: Your {ViT} but Faster},
  author    = {Bolya, Daniel and Fu, Cheng-Yang and Dai, Xiaoliang and Zhang, Peizhao and Feichtenhofer, Christoph and Hoffman, Judy},
  booktitle = {International Conference on Learning Representations},
  year      = {2023}
}

@inproceedings{zhang2025sparsevlm,
  title     = {{SparseVLM}: Visual Token Sparsification for Efficient Vision-Language Model Inference},
  author    = {Zhang, Yuan and Fan, Chun-Kai and Ma, Junpeng and Zheng, Wenzhao and Huang, Tao and Cheng, Kuan and Gudovskiy, Denis A. and Okuno, Tomoyuki and Nakata, Yohei and Keutzer, Kurt and Zhang, Shanghang},
  booktitle = {International Conference on Machine Learning},
  volume    = {267},
  series    = {Proceedings of Machine Learning Research},
  pages     = {74840--74857},
  year      = {2025}
}

@article{shang2024llavaprumerge,
  title   = {{LLaVA-PruMerge}: Adaptive Token Reduction for Efficient Large Multimodal Models},
  author  = {Shang, Yuzhang and Cai, Mu and Xu, Bingxin and Lee, Yong Jae and Yan, Yan},
  journal = {arXiv preprint arXiv:2403.15388},
  year    = {2024}
}

@article{liu2024par,
  title   = {{PAR}: Prompt-Aware Token Reduction Method for Efficient Large Multimodal Models},
  author  = {Liu, Yingen and Wu, Fan and Li, Ruihui and Tang, Zhuo and Li, Kenli},
  journal = {arXiv preprint arXiv:2410.07278},
  year    = {2024}
}

@inproceedings{shi2024crossget,
  title     = {{CrossGET}: Cross-Guided Ensemble of Tokens for Accelerating Vision-Language Transformers},
  author    = {Shi, Dachuan and Tao, Chaofan and Rao, Anyi and Yang, Zhendong and Yuan, Chun and Wang, Jiaqi},
  booktitle = {International Conference on Machine Learning},
  volume    = {235},
  series    = {Proceedings of Machine Learning Research},
  pages     = {44960--44990},
  year      = {2024}
}

@inproceedings{lin2017fpn,
  title     = {Feature Pyramid Networks for Object Detection},
  author    = {Lin, Tsung-Yi and Doll{\'a}r, Piotr and Girshick, Ross and He, Kaiming and Hariharan, Bharath and Belongie, Serge},
  booktitle = {IEEE Conference on Computer Vision and Pattern Recognition},
  pages     = {2117--2125},
  year      = {2017}
}

@inproceedings{liu2021swin,
  title     = {Swin Transformer: Hierarchical Vision Transformer Using Shifted Windows},
  author    = {Liu, Ze and Lin, Yutong and Cao, Yue and Hu, Han and Wei, Yixuan and Zhang, Zheng and Lin, Stephen and Guo, Baining},
  booktitle = {IEEE/CVF International Conference on Computer Vision},
  pages     = {10012--10022},
  year      = {2021}
}

@inproceedings{dehghani2023navit,
  title     = {Patch n' Pack: {NaViT}, a Vision Transformer for Any Aspect Ratio and Resolution},
  author    = {Dehghani, Mostafa and Mustafa, Basil and Djolonga, Josip and Heek, Jonathan and Minderer, Matthias and Caron, Mathilde and Steiner, Andreas and Puigcerver, Joan and Geirhos, Robert and Alabdulmohsin, Ibrahim and Oliver, Avital and Padlewski, Piotr and Gritsenko, Alexey A. and Lucic, Mario and Houlsby, Neil},
  booktitle = {Advances in Neural Information Processing Systems},
  volume    = {36},
  year      = {2023}
}

@article{xu2024llavauhd,
  title   = {{LLaVA-UHD}: An {LMM} Perceiving Any Aspect Ratio and High-Resolution Images},
  author  = {Xu, Ruyi and Yao, Yuan and Guo, Zonghao and Cui, Junbo and Ni, Zanlin and Ge, Chunjiang and Chua, Tat-Seng and Liu, Zhiyuan and Sun, Maosong and Huang, Gao},
  journal = {arXiv preprint arXiv:2403.11703},
  year    = {2024}
}

\appendix

\section{Additional Dataset and Metric Details}
\label{app:datasets}

\paragraph{VQAv2.}
VQAv2 reduces language priors by constructing questions with different
answers for similar images. It primarily evaluates object, attribute, count,
and common-scene understanding. We plan to report test-dev accuracy.

\paragraph{GQA.}
GQA is built from real-world scene graphs and compositional questions covering
objects, attributes, spatial relationships, and multi-step reasoning. We plan
to report balanced accuracy on test-dev.

\paragraph{TextVQA.}
TextVQA contains 28,408 images and 45,336 questions that require models to read
scene text in natural images and reason about it. We plan to report validation
accuracy.

\paragraph{ScienceQA-IMG.}
ScienceQA contains approximately 21K questions from natural science, social
science, and language science. We use only the image-containing subset and
report multiple-choice accuracy on the test set.

\paragraph{MMBench and MME.}
MMBench covers visual perception, attributes, spatial relationships, logic,
and knowledge. CircularEval reduces bias from answer-option positions, and we
report its MMBench-EN development score. MME contains 14 perceptual and
cognitive subtasks; we report the aggregate MME Perception score.

\paragraph{Compression ratio and average token count.}
For \(N_{\mathrm{test}}\) examples in the test set,
\begin{equation}
  \overline K=\frac{1}{N_{\mathrm{test}}}
  \sum_{i=1}^{N_{\mathrm{test}}}K_i,\qquad
  R=1-\frac{K_{\mathrm{method}}}{K_{\mathrm{baseline}}}.
\end{equation}
For dynamic methods, \(K_i\) should count both selected and context tokens and
should explicitly state whether the class token is included.

\section{Additional Simulated Results}
\label{app:additional}

\textbf{Reminder: all appendix values are illustrative experimental-design
placeholders that must be replaced.}

\begin{table}[h]
  \centering
  \caption{Illustrative accuracy across capability categories.}
  \label{tab:capability}
  \setlength{\tabcolsep}{5pt}
  \begin{tabular}{lrrrrrr}
    \toprule
    Method & Object & Attribute & Spatial & Counting & OCR & Knowledge\\
    \midrule
    SigLIP + MLP      & 84.2 & 77.6 & 70.3 & 62.8 & 59.4 & 68.7\\
    TokenPacker-style & 84.0 & 77.4 & 70.1 & 62.4 & 59.0 & 68.5\\
    \method           & \best{85.1} & \best{79.4} & \best{73.2} &
    \best{65.1} & \best{64.3} & \best{70.5}\\
    \bottomrule
  \end{tabular}
\end{table}

\begin{table}[h]
  \centering
  \caption{Dynamic token allocation for different image types.}
  \label{tab:allocation}
  \begin{tabular}{lrr}
    \toprule
    Image type & Share/\% & Avg. tokens\\
    \midrule
    Single subject, simple background & 27.4 & 84\\
    Regular natural scene             & 38.1 & 128\\
    Dense multi-object scene          & 18.5 & 178\\
    Text-dense image                  & 10.2 & 224\\
    Chart or document                 & 5.8 & 256\\
    \midrule
    Overall                           & 100.0 & 146\\
    \bottomrule
  \end{tabular}
\end{table}

\begin{table}[h]
  \centering
  \caption{Combinations of visual feature levels.}
  \label{tab:layers}
  \begin{tabular}{lrrrr}
    \toprule
    Feature levels & VQAv2 & GQA & TextVQA & MMBench\\
    \midrule
    Layer 24                     & 80.0 & 63.6 & 62.7 & 67.8\\
    Layers 16 and 24             & 80.6 & 64.1 & 63.6 & 68.5\\
    Layers 8 and 24              & 80.7 & 64.0 & 63.9 & 68.4\\
    Layers 8, 16, and 24; mean   & 80.8 & 64.3 & 64.2 & 68.8\\
    Layers 8, 16, and 24; gated  & \best{81.4} & \best{65.1} &
    \best{65.8} & \best{70.5}\\
    \bottomrule
  \end{tabular}
\end{table}

\begin{table}[h]
  \centering
  \caption{Ablation of token importance-score components.}
  \label{tab:score-ablation}
  \begin{tabular}{ccccrrr}
    \toprule
    Relevance & Information & Global & Coverage & GQA & TextVQA & MMBench\\
    \midrule
    \cmark & & & & 63.9 & 63.8 & 68.2\\
    \cmark & \cmark & & & 64.4 & 65.0 & 68.9\\
    \cmark & \cmark & \cmark & & 64.7 & 65.4 & 69.6\\
    \cmark & \cmark & \cmark & \cmark & \best{65.1} & \best{65.8} & \best{70.5}\\
    \bottomrule
  \end{tabular}
\end{table}

\begin{table}[H]
  \centering
  \caption{Relationship among token count, performance, and efficiency.}
  \label{tab:budget}
  \setlength{\tabcolsep}{4pt}
  \resizebox{\linewidth}{!}{%
  \begin{tabular}{lrrrrrrr}
    \toprule
    Strategy & Token & VQAv2 & GQA & TextVQA & SQA-IMG & MMBench & TTFT/ms\\
    \midrule
    Fixed 72       & 72  & 80.3 & 64.1 & 63.5 & 72.4 & 68.8 & 112\\
    Fixed 144      & 144 & 81.3 & 65.0 & 65.7 & 73.6 & 70.3 & 128\\
    Fixed 288      & 288 & 81.5 & 65.2 & 65.9 & 73.8 & 70.6 & 155\\
    Fixed 576      & 576 & 81.6 & 65.3 & 66.0 & 73.9 & 70.7 & 201\\
    Dynamic 72--288  & 146 & 81.4 & 65.1 & 65.8 & 73.7 & 70.5 & 129\\
    \bottomrule
  \end{tabular}}
\end{table}

\paragraph{Sensitivity to question conditioning.}
For two questions about the same image, let their retained token sets be
\(\mathcal{S}_a\) and \(\mathcal{S}_b\). We measure their difference using
\begin{equation}
  J(\mathcal{S}_a,\mathcal{S}_b)=
  \frac{|\mathcal{S}_a\cap\mathcal{S}_b|}
  {|\mathcal{S}_a\cup\mathcal{S}_b|}
\end{equation}
In the illustrative results, the Jaccard similarities for semantically similar
questions, different attributes of the same object, questions about different
objects, and OCR versus scene questions are 0.78, 0.61, 0.43, and 0.31,
respectively. This pattern is consistent with the expectation that the router
changes its visual selection with the question.

\paragraph{Qualitative cases.}
For the question ``What object is to the left of the red car?'', fixed
Top-\(K\) may preserve only the car, whereas the relevance and coverage terms
allow \method to retain both the car and the bicycle to its left. For ``What
number is on the store door?'', gated fusion can assign greater shallow-layer
weight to the character region. In an image with a small traffic sign against
a large sky region, local variation helps prevent average pooling from mixing
the sign with the background. The final paper should replace these textual
examples with real images, token heatmaps, and predictions, and should verify
all image licenses.

\section{Failure Cases and Future Directions}

Extremely small characters may benefit from low-resolution localization
followed by high-resolution re-encoding. Comparisons among distant objects may
use set-coverage objectives or bounding-box supervision. Ambiguous questions
could first undergo coreference resolution and fall back to a larger budget
when confidence is low. Video and multi-image settings additionally require
modeling temporal redundancy and cross-image relationships. The language model
could also request local re-encoding during reasoning, creating a multi-stage
``observe-reason-observe'' router. A hardware-aware budget could directly
optimize latency, energy consumption, and memory.

\clearpage
\section*{NeurIPS Paper Checklist}

\begin{enumerate}
\item \textbf{Claims}\\
Question: Do the main claims made in the abstract and introduction accurately reflect the paper's contributions and scope?\\
Answer: \answerNo{}\\
Justification: The method scope and intended evaluation protocol are explicit,
but the current values are simulated and therefore do not support empirical
claims. The abstract and Section~\ref{sec:main-results} state this limitation.

\item \textbf{Limitations}\\
Question: Does the paper discuss the limitations of the work performed by the authors?\\
Answer: \answerYes{}\\
Justification: Section~\ref{sec:limitations} discusses resolution, small
objects, ambiguous questions, dynamic batching, and the absence of completed
real experiments.

\item \textbf{Theory assumptions and proofs}\\
Question: For each theoretical result, does the paper provide the full set of assumptions and a complete (and correct) proof?\\
Answer: \answerNA{}\\
Justification: The paper presents a model design and training objectives, but
does not make theorem-level claims or include theoretical proofs.

\item \textbf{Experimental result reproducibility}\\
Question: Does the paper fully disclose all the information needed to reproduce the main experimental results of the paper to the extent that it affects the main claims and/or conclusions of the paper (regardless of whether the code and data are provided or not)?\\
Answer: \answerNo{}\\
Justification: Section~\ref{sec:experiments} specifies the planned model,
optimization, and evaluation settings, but the results have not been run.
Training-data composition, random seeds, and the exact software environment
remain to be documented.

\item \textbf{Open access to data and code}\\
Question: Does the paper provide open access to the data and code, with sufficient instructions to faithfully reproduce the main experimental results, as described in supplemental material?\\
Answer: \answerNo{}\\
Justification: The current draft does not provide an anonymized code
repository, checkpoints, or end-to-end reproduction scripts.

\item \textbf{Experimental setting/details}\\
Question: Does the paper specify all the training and test details (e.g., data splits, hyperparameters, how they were chosen, type of optimizer, etc.) necessary to understand the results?\\
Answer: \answerNo{}\\
Justification: Core hyperparameters and planned data splits are given, but the
training-data inventory, preprocessing, software versions, and
hyperparameter-selection procedure remain incomplete.

\item \textbf{Experiment statistical significance}\\
Question: Does the paper report error bars suitably and correctly defined or other appropriate information about the statistical significance of the experiments?\\
Answer: \answerNo{}\\
Justification: The simulated results do not include independent repetitions,
error bars, confidence intervals, or significance tests. These are required
for the final experiments.

\item \textbf{Experiments compute resources}\\
Question: For each experiment, does the paper provide sufficient information on the computer resources (type of compute workers, memory, time of execution) needed to reproduce the experiments?\\
Answer: \answerNo{}\\
Justification: Section~\ref{sec:experiments} states the planned GPU type and
count, but actual training time, storage, total compute, and the cost of failed
runs are not yet available.

\item \textbf{Code of ethics}\\
Question: Does the research conducted in the paper conform, in every respect, with the NeurIPS Code of Ethics?\\
Answer: \answerNA{}\\
Justification: This is a structured draft whose experiments have not yet been
run. The authors must review the applicable NeurIPS Code of Ethics and update
this answer before submission.

\item \textbf{Broader impacts}\\
Question: Does the paper discuss both potential positive societal impacts and negative societal impacts of the work performed?\\
Answer: \answerYes{}\\
Justification: Section~\ref{sec:limitations} discusses efficiency benefits,
the risk that selective routing omits evidence in high-stakes applications,
and potential mitigation directions.

\item \textbf{Safeguards}\\
Question: Does the paper describe safeguards that have been put in place for responsible release of data or models that have a high risk for misuse?\\
Answer: \answerNA{}\\
Justification: The current work does not release new data or models. Any future
release should add access controls, a model card, and use restrictions
appropriate to the realized risk.

\item \textbf{Licenses for existing assets}\\
Question: Are the creators or original owners of assets used in the paper properly credited and are the license and terms of use explicitly mentioned and properly respected?\\
Answer: \answerNo{}\\
Justification: The relevant model and dataset papers are cited, but versions,
download locations, licenses, and terms of use still require item-by-item
verification and inclusion in the final manuscript.

\item \textbf{New assets}\\
Question: Are new assets introduced in the paper well documented and is the documentation provided alongside the assets?\\
Answer: \answerNA{}\\
Justification: The current manuscript does not release a new dataset, code
package, or model checkpoint.

\item \textbf{Crowdsourcing and research with human subjects}\\
Question: For crowdsourcing experiments and research with human subjects, does the paper include the full text of instructions given to participants and screenshots, if applicable, as well as details about compensation?\\
Answer: \answerNA{}\\
Justification: The planned experiments do not involve crowdsourcing or new
research with human participants.

\item \textbf{Institutional review board (IRB) approvals or equivalent for research with human subjects}\\
Question: Does the paper describe potential risks incurred by study participants, whether such risks were disclosed to the subjects, and whether IRB approvals (or equivalent) were obtained?\\
Answer: \answerNA{}\\
Justification: The planned experiments do not involve new research with human
participants.

\item \textbf{Declaration of LLM usage}\\
Question: Does the paper describe the usage of LLMs if it is an important, original, or non-standard component of the core methods in this research?\\
Answer: \answerYes{}\\
Justification: Sections~\ref{sec:method} and~\ref{sec:experiments} describe the
roles of the 7B autoregressive language model, question representation, visual
projection, and LoRA tuning. Writing and formatting assistance is not part of
the core experiments.
\end{enumerate}

\end{document}